\documentclass[11pt]{article}

\usepackage[final]{acl}

\usepackage{times}
\usepackage{latexsym}

\usepackage[T1]{fontenc}

\usepackage[utf8]{inputenc}

\usepackage{microtype}

\usepackage{inconsolata}

\usepackage{graphicx}

\usepackage{amsmath}
\usepackage{amssymb}
\usepackage{fancyvrb}
\usepackage{fvextra}
\usepackage{xurl}

\newcommand{\cost}{\text{cost}}
\newcommand{\early}{\text{early}}
\newcommand{\phiGlob}{\Phi}
\newcommand{\phiLoc}{\Phi^{(k)}}

\title{CaRL-EM: Cost-Aware Reinforcement Learning for Entity Matching with LLMs}

\author{Chaohui Guo, Michel Klein, Zhisheng Huang \\
       Department of Computer Science, Vrije Universiteit Amsterdam, Netherlands \\
         \texttt{c.guo2@vu.nl} }

\begin{document}
\maketitle
\begin{abstract}
Entity matching (EM) requires fine-grained contextual understanding and domain knowledge. Recent work shows that large language models (LLMs) can serve as strong matchers across domains, but most methods either make independent pairwise decisions or rely on manually designed composite pipelines, thus lacking flexibility in realistic multi-candidate settings. At the same time, they typically ignore inference cost at scale. We formulate LLM-based EM with candidates as a cost-aware sequential decision problem and propose CaRL-EM, a reinforcement learning controller that manages LLM operations. Given the state of an anchor record, its candidate set, and the cost, CaRL-EM adaptively chooses among different operators (\textsc{Match} / \textsc{Compare} / \textsc{Select} / \textsc{Decide}) and model capacities to maximize a quality–cost objective. The policy interacts with abstract operators, allowing the same controller to be reused with different underlying LLM backends at inference time without retraining. Experiments on 7 benchmarks show that CaRL-EM (i) learns to dynamically plan the usage of inexpensive and expensive operators based on task complexity, (ii) achieves robust zero-shot transfer across diverse datasets and domains, and (iii) consistently achieves a better quality–cost trade-off than strong LLM-based baselines and manually designed pipelines, yielding a lower inference cost at comparable or higher quality.

\end{abstract}

\section{Introduction}

\begin{figure}[t]
  \includegraphics[width=\columnwidth]{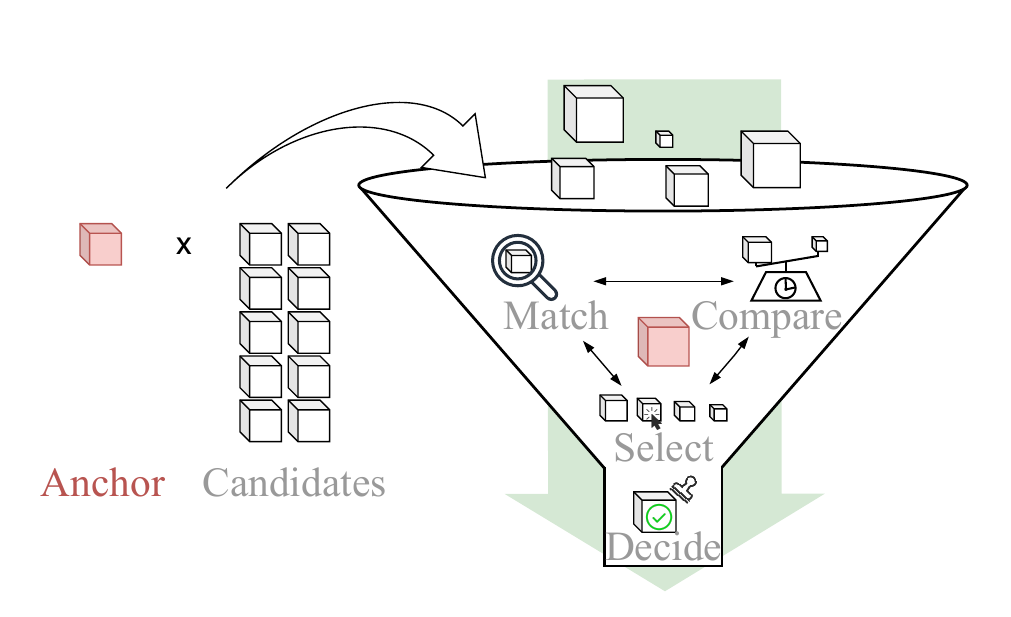}
  \caption{\textbf{Overview of CaRL-EM.}
    The RL agent ranks the candidates by sequentially choosing between tools \textsc{Match}, \textsc{Compare} and \textsc{Select}.
}
  \label{fig:teaser}
\end{figure}

Entity matching (EM) is a core component of entity resolution pipelines, supporting data integration, knowledge base construction, and downstream analytics in multiple domains~\citep{shahbazi2023fairness}. Given an anchor record and a set of candidates retrieved by blocking, an EM system must identify which candidates refer to the same real-world entity~\citep{papadakis2020blocking, thirumuruganathan2021deepblocking, paulsen2023sparkly}.

Recent work shows that large language models (LLMs)
can rival supervised deep models in pairwise matching, even under zero- or few-shot prompting
~\citep{peeters2023using,li2024leveraging}.
However, most LLM-based EM methods focus on whether an anchor matches each candidate in isolation, ignoring mutual interactions among candidates within the same candidate set and the set-level exclusivity constraint that, in clean-clean EM, at most one candidate should be selected.
In addition, EM at an industrial scale involves millions of anchors and large candidate sets, making cost a major concern~\citep{Konda2016Magellan}. This becomes more significant with the application of LLMs.

To address these issues,
\textsc{COMEM}~\citep{wang2025match} use a pipeline to allow more interaction among candidates. While these pipelines improve quality, they remain treating cost as a fixed byproduct of the architecture, applying the same sequence of operations regardless of an instance’s difficulty~\citep{chen2023frugalgpt,ong2025routellm}.

In this paper, we formulate LLM-based multi-candidate EM with blocking as a cost-aware sequential decision problem and propose CaRL-EM, as shown in Figure~\ref{fig:teaser}, a reinforcement learning (RL) controller that manages LLM-based EM operations.
We focus on the standard clean-clean setting, in which each anchor has at most one true match in its retrieved candidate set; thus, the task studied here is to select one match or \textsc{None} from the candidate set.
For each anchor and its candidate set, CaRL-EM maintains a compact state. We detail the state representation in \S3.2. At each step, the controller selects a high level operator, \textsc{Match}, \textsc{Compare}, \textsc{Select}, or \textsc{Decide}, and a model capacity, thereby deciding whether to perform low-cost local refinement, more expensive listwise selection over the current shortlist of candidates, or terminate and output a prediction. The policy is trained with cost-aware rewards under an abstract two level cost model, more details in \S3.1.
CaRL-EM treats these operators as black box actions annotated with abstract cost and is separated from the underlying LLMs, so that once the policy has been trained, stronger LLMs can be plugged in at test time without retraining the controller. In summary, our work makes the following contributions:
\begin{itemize}

\item\textbf{Problem formulation.} We formulate LLM-based multi-candidate EM with blocking as a cost-aware sequential decision problem under RL, where a policy observes the matching state and accumulated cost information, and decides which operator to apply next. To the best of our knowledge, this is the first work to cast LLM-driven EM into a cost-aware sequential RL process.

\item\textbf{CaRL-EM.} We propose CaRL-EM, a cost-aware RL controller that combines \textsc{Match} / \textsc{Compare} / \textsc{Select} / \textsc{Decide} operators and two level abstract cost models to reduce unnecessary calls. The controller is independent of specific LLMs, so backends can be swapped at test time without retraining.

\item\textbf{High efficiency and performance.} On 7 benchmarks spanning products, citations, and movies, CaRL-EM outperforms the best manually designed composite pipelines,
achieves higher F1 score,
and \textbf{reduces 78\%} of the cost
This yields a better quality cost trade-off. Compared to domain-specific supervised EM models, CaRL-EM attains approximately 89\% of their performance without any fine-tuning, while
\textbf{reducing 94\%} of the expense.

\end{itemize}

\section{Related Work}
\subsection{Traditional and Pretrained EM}
Early EM mainly used string similarity and manual rules ~\citep{papadakis2020blocking,barlaug2020survey}. While neural models such as DeepER and DeepMatcher improved semantic capture, they remain heavily dependent on labeled data ~\citep{ebraheem2017deeper,mudgal2018deep}. Recent pretrained language models further leverage cross-encoders like \textsc{Ditto} ~\citep{li2020deep} or dual-encoders ~\citep{shah2018extreme,tracz2020bert}.
However, these methods typically treat $(anchor, candidate)$ pairs independently, ignoring mutual interactions and global consistency. Moreover, their reliance on fine-tuning limits their transferability to new domains ~\citep{li2020deep,peeters2021dualobjective}.

\subsection{LLM-Based Strategies}

LLMs can perform pairwise matching in a zero- or few-shot manner with performance often rivaling supervised models, while providing explanations of their decisions \cite{Ralph2025Entity}. Most methods prompt the LLM to output a Yes/No decision for each pair, which keeps the pairwise limitation.

To go beyond independent decisions, \textsc{COMEM} analyzes three LLM-driven strategies for multi-candidate EM: \textsc{Match},
\textsc{Compare},
and \textsc{Select}
\cite{wang2025match}. It manually builds a pipeline that uses cheap steps to re-rank candidates and then calls a stronger LLM to select the final match. However, the pipeline structure is the same for all anchors, and the inference cost is implicitly determined by the chosen pipeline.

\subsection{Cost-Aware Control with RL}
LLMs often work well without much tuning, but they can be costly to run at scale. This has led to growing interest in controlling LLM inference under resource constraints. Recent work uses lightweight policies or RL to decide when to call external tools or how much context to use, trading off accuracy for token cost~\citep{feng2025retool,zhang2024treacle}. These approaches are mainly studied in question answering or reasoning tasks rather than EM. In contrast, EM tasks that require large-scale LLM calls typically rely on fixed pipelines.

We cast multi candidate EM as a cost-aware sequential decision problem.
CaRL-EM learns to chooses \textsc{Match}, \textsc{Compare}, \textsc{Select}, and \textsc{Decide},
as well as LLMs with different costs, to navigate the accuracy and cost trade-off.

\section{CaRL-EM: A Cost-Aware RL Controller for LLM-Based EM}
We consider the standard blocked EM setting. For each anchor record $a$, a blocking retrieves a candidate set $C(a) = \{c_1, \dots, c_n\}$. The goal is to decide which candidate matches $a$, or output \textsc{None}.  We focus on the common clean-clean scenario in which at most one candidate in $C(a)$ is a true match~\citep{gemmell2011improving}.
We cast multi-candidate EM for each anchor as a cost-aware sequential decision problem and learn a policy that chooses which action to apply next, balancing both expected matching quality and inference cost. All hyperparameters introduced in this section, including $\rho$, $\lambda$, $\eta$, $\sigma$, $\gamma$, and top-$k$, are specified with their values in Appendix~\ref{appendix:A}.

\subsection{LLM-Based Operators and Abstract Cost}

CaRL-EM uses four high-level actions: three LLM-based operators (\textsc{Match}, \textsc{Compare}, \textsc{Select}) and a terminal action (\textsc{Decide}), as shown in Figure~\ref{fig:operators}. The LLM-based operators are implemented with dedicated prompting templates in Appendix~\ref{appendix:B}.

\noindent\textbf{\textsc{Match.}}
Given an anchor candidate pair $(a, c_i)$, a \textsc{Match} call asks a lightweight matcher, Flan-T5-xl~\citep{Raffel2020Exploring} is used for \textsc{Match} and \textsc{Compare} in our experiment, to output YES/NO. We use the model probability of emitting YES to update the internal confidence score.

\noindent\textbf{\textsc{Compare.}}
A \textsc{Compare} call presents $a$ with two candidates $(c_i, c_j)$ and asks which candidate is more likely to match $a$; the resulting preference is translated into a small update of their respective scores $s_i$ and $s_j$, sharpening the score distribution.

\noindent\textbf{\textsc{Select.}}
A \textsc{Select} call presents $a$ and a shortlist $\tilde{C}(a)$ of top-$k$ candidates ranked by the current scores. It prompts the LLM to perform listwise reasoning to identify the most likely match within this group or output None. The selected candidate receives a score boost, while others in the shortlist are penalized. This listwise feedback refines the global ranking without making a final decision.

\noindent\textbf{\textsc{Decide.}}
\textsc{Decide} is the only terminal operator. The action set includes $\{\textsc{Decide}(i)\}_{i=1}^n \cup \{\textsc{Decide}(\mathrm{None})\}$. The policy directly chooses a candidate or None without a specific thresholding.

\begin{figure}[t]
  \includegraphics[width=\columnwidth]{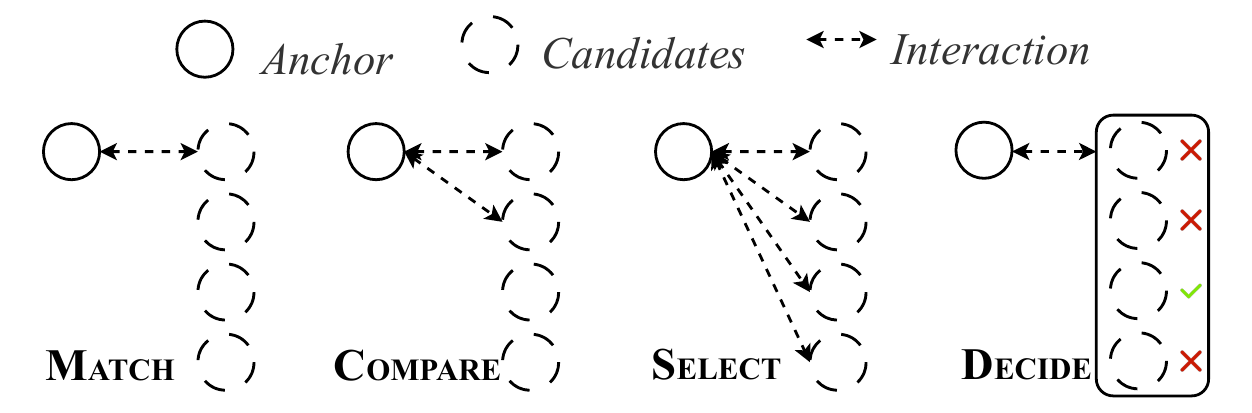}
  \caption{Interaction patterns of the operators used in CaRL-EM: \textsc{\textbf{Match}}, \textsc{\textbf{Compare}}, \textsc{\textbf{Select}}, and \textsc{\textbf{Decide}}.
  }
  \label{fig:operators}
\end{figure}

Operators vary significantly in complexity and input scope, often necessitating different model capacities. Local operations like \textsc{Match} and \textsc{Compare} are computationally lighter and can be handled by smaller models, whereas the listwise \textsc{Select} operator requires reasoning over a larger context of multiple candidates, often demanding a larger, more capable model \cite{wang2025match}. This disparity in both model size and prompt length leads to distinct inference costs.

We capture this difference using a two level abstract cost model. Each operator is assigned a cost label in $\{\text{low}, \text{high}\}$, which we normalize to numerical values, e.g., $\kappa_\ell$ for low-cost local \textsc{Match}/\textsc{Compare} calls and $\kappa_h=\rho \kappa_\ell$ for high-cost shortlist \textsc{Select} calls. We set $\rho{=}2.5$ by default. Increasing it to $\rho{=}5$ yields nearly the same $F1_{\text{macro}}$, but increases \textsc{Match}/\textsc{Compare} usage and leads to higher cross-benchmark variance; more details are provided in Appendix~\ref{appendix:C}.

These numbers are not tied to any particular pricing scheme and are intended to reflect relative expense in terms of tokens and API calls. Thus, the abstract cost is a control signal for policy learning rather than an exact accounting model of any single deployment environment. This abstraction decouples the learned policy from a specific model. By training on relative cost tiers, CaRL-EM allows users to swap different underlying LLMs for the operator at inference time without retraining the controller. Given a sequence of actions $\alpha_{1:T}$ for an anchor, the total abstract cost is
\begin{equation*}
    \mathcal{C}(\alpha_{1:T}) = \sum_{t=1}^{T} \cost(\alpha_t),
\label{eq:total_cost}
\end{equation*}

and this cost enters the RL objective as a penalty, encouraging the policy to use cheap operators whenever they suffice.

\subsection{MDP Formulation}
Actions change the candidate state and incur cost, so decisions are sequential, including when to stop with \textsc{Decide}.
We therefore model the decision process for each anchor as an episodic Markov decision process (MDP)~\citep{sutton1998reinforcement}.
At each step, an action invokes an operator that updates the state, and we learn a policy $\pi$ to maximize the expected return $\mathbb{E}_{\pi}\!\left[\sum_{t=1}^{T} r_t\right]$ until \textsc{Decide} or $T_{\max}$.

\noindent\textbf{State.}
At step $t$, the policy observes a state vector $\mathbf{x}_t$ that captures the current belief state, resource consumption, and interaction history. Formally, let $N$ be the maximum number of candidates and $H$ be the length of the action history window. The state vector is constructed as the concatenation of several feature groups:
\begin{equation*}
    \mathbf{x}_t = [ \mathbf{v}_{\text{score}} \oplus \mathbf{v}_{\text{mask}} \oplus \mathbf{v}_{\text{freq}} \oplus \mathbf{v}_{\text{global}} \oplus \mathbf{v}_{\text{hist}} (\oplus \mathbf{v}_{\text{emb}}) ].
\end{equation*}

$\mathbf{v}_{\text{score}}$ contains the current confidence scores, $\mathbf{v}_{\text{mask}}$ indicates whether each candidate is still eligible to be processed by a \textsc{Match} operator (e.g., has not exceeded a per-candidate call limit), and $\mathbf{v}_{\text{freq}}$ tracks the frequency of each candidate in \textsc{Compare} calls. The global context $\mathbf{v}_{\text{global}}$
consists of the accumulated abstract cost, the normalized time step $t/T_{\max}$, and the current maximum candidate score. To detect loops or repetitive patterns, $\mathbf{v}_{\text{hist}}$
encodes the last $H$ high level actions as a one-hot vector. Optionally, dense semantic embeddings of the anchor and the current top-scoring candidate ($\mathbf{v}_{\text{emb}}$) are appended to provide grounding.
During training, we add small Gaussian noise $\sigma$ into $\mathbf{x}_t$ to enhance policy robustness.

\noindent\textbf{Action.}
At step $t$, the controller chooses a discrete action $\alpha_t$ that specifies an operator and its operands. Actions include $\textsc{Match}(i)$ for inspecting candidate $i$, $\textsc{Compare}(j)$ for evaluating a selected champion--challenger pair, $\textsc{Select}(k)$ for listwise selection over the current top-$k$ candidates, and terminal $\textsc{Decide}(i)$ / $\textsc{Decide}(\text{None})$. The episode ends when \textsc{Decide} is chosen or when $T_{\max}$ is reached. Executing $\alpha_t$ invokes the corresponding LLM operator, updates candidate scores and usage statistics, and accumulates abstract cost. The full equations of candidate score updating are provided in Appendix~\ref{appendix:D}.

\noindent\textbf{Reward.}
We design the reward to favor correct final decisions while keeping the cost low.
Let $s_i^{(t)}$ be the score of candidate $i$ at step $t$, $y_i\in\{0,1\}$ its label, and $S_t$ the active candidates.
We define a global margin $\phiGlob_t$ as the score gap between the best true match and the best non-match; if no match exists, it is the negative best score:

\begin{equation*}
\phiGlob_t =
\begin{cases}
\displaystyle \max_{i: y_i = 1} s_i^{(t)} - \max_{j: y_j = 0} s_j^{(t)}, & \text{if } \exists i \in S_t, y_i=1,\\[8pt]
\displaystyle -\max_{j \in S_t} s_j^{(t)}, & \text{otherwise},
\end{cases}
\end{equation*}

At a terminal step $T$, the agent outputs $\hat y_T \in S_T \cup \{\varnothing\}$ and receives a correctness reward:

\begin{equation*}
R_{\text{term}}(\hat y_T, y) =
\begin{cases}
R_{\text{sel,cor}}, & \hat y_T = i, y_i = 1,\\
R_{\text{sel,wro}}, & \hat y_T = i, y_i = 0,\\
R_{\text{none,cor}},& \hat y_T = \varnothing, \forall i, y_i = 0,\\
R_{\text{none,wro}},& \hat y_T = \varnothing, \exists i, y_i = 1.
\end{cases}
\end{equation*}

The total terminal reward is formulated as:
\begin{equation*}
\begin{split}
r_T = R_{\text{term}} + r^{\text{eff}}_T
-\lambda\,\cost(\alpha_T) -\beta\,\early_T +\epsilon_T,
\end{split}
\end{equation*}

where $r^{\text{eff}}_T$ rewards stopping before the deadline $T_{\max}$, $\cost(\alpha_t)$ reflects the abstract cost, and $\early_T$ penalizes premature decisions with low confidence.

For non-terminal steps $t < T$, we use shaping to guide the agent~\citep{ng1999policy,wiewiora2003potential}:
\begin{equation*}
\begin{split}
r_t &= -\lambda\,\cost(\alpha_t) +\eta(\gamma\,\phiGlob_{t+1}-\phiGlob_t) \\
    &\quad +\eta_k\,\mathbb{I}[\alpha_t = \textsc{Select}] (\gamma\,\phiLoc_{t+1}-\phiLoc_t) +\epsilon_t.
\end{split}
\end{equation*}

\subsection{Policy, Training, and Inference}
CaRL-EM uses a lightweight policy network $\pi_\theta(\alpha_t \mid \mathbf{x}_t)$, implemented as a 3-layer MLP (768--384--192), which maps the state vector $\mathbf{x}_t$ to a distribution over high level actions. We train $\pi_\theta$ with Proximal Policy Optimization (PPO) and a learned value baseline~\citep{schulman2017proximal}.

At test time,
the controller starts from the initial state
and iteratively selects actions until it chooses \textsc{Decide} or reaches the step limit $T_{\max}$. Because the controller interacts only with abstract operators and their cost labels, and is decoupled from the underlying LLMs that implement them, stronger LLMs can be plugged in at inference time without retraining the policy, as long as the operator interfaces and relative cost levels are preserved. This allows CaRL-EM to benefit from future improvements in LLM capabilities while retaining the learned decision strategy.

\section{Experiments}
\paragraph{LLMs used in our experiments.}

We evaluate CaRL-EM with a diverse set of LLM backends, covering both proprietary commercial APIs~\citep{team2024gpt} and open weight models~\citep{chung2024scaling,grattafiori2024llama,agarwal2025gpt,ernie2025technicalreport,team2025gemma,yang2025qwen3}; more detail are in Table~\ref{tab:models}.
For open weight models,
we report a price range collected from several major inference providers\footnote{
Prices are taken from providers: DeepInfra \url{https://deepinfra.com/}; Fireworks AI \url{https://fireworks.ai/}; Together.ai \url{https://www.together.ai/}; Novita \url{https://novita.ai/}; Groq \url{https://groq.com/}}.

\begin{table}[t]
    \centering
    \small
    \begin{tabular}{l l c l l}
    \hline
    \textbf{Model Name} & \textbf{Size} & \textbf{CoT} & \multicolumn{2}{c}{\textbf{Cost (\$/1M tokens)}} \\
    \cline{4-5}
     & & & \textbf{Input} & \textbf{Output} \\
    \hline
    Flan-T5-xl        & 3B  & --          & 0.10        & 0.10        \\
    Llama-3.1 & 8B  & --          & $0.03_{\sim0.22}$        & $0.03_{\sim0.22}$        \\
    GPT-oss & 20B & \checkmark  & $0.05_{\sim0.10}$ & $0.20_{\sim0.50}$        \\
    ERNIE 4.5         & 21B & \checkmark  & 0.07        & 0.28        \\
    Gemma 3         & 27B & --  & 0.1      & 0.2        \\
    Qwen3 & 30B & \checkmark & $0.20_{\sim0.89}$        & $0.20_{\sim7.90}$        \\
    GPT-4o mini       & --  & --          & 0.15        & 0.60        \\
    \hline
    \end{tabular}
    \caption{
      Model specifications and costs. CoT: Chain-of-Thought support. For open weight models, we report the price range from multiple providers.
    }
    \label{tab:models}
\end{table}

\noindent\textbf{Datasets and Zero-shot Transfer Setting.}
We train our approach on Abt-buy (AB) and evaluate it on standard entity resolution benchmarks extensively used in prior literature~\citep{mudgal2018deep}. These datasets cover diverse domains, including e-commerce, academic citations, movies, and restaurants. Following the experimental protocol established in \textsc{COMEM}~\citep{wang2025match}, we employ a blocking stage using Sparkly to retrieve the top-10 most likely candidates from the target table for each anchor record. Blocking recall@10 is high across all datasets, ranging from 94.89\%-to 99.96\%, so remaining errors mainly reflect the quality of decisions. The final evaluation set contains 400 anchors per dataset, including 300 anchors with one true match, and the rest have no true match. This setup shifts the task from isolated pairwise classification to a realistic \textit{multi-candidate selection} problem in the clean-clean setting, where the model must choose one candidate or reject all candidates.

To evaluate the generalization capability of CaRL-EM, we use a zero-shot transfer protocol. CaRL-EM is trained only once on the AB dataset and evaluates it on the other seven benchmarks without any fine-tuning or adaptation. Since AB comes from a different domain and source than the test datasets, the controller cannot learn target-domain knowledge during training, and there is no data leakage. In contrast, for supervised baselines, like \textsc{Ditto}~\citep{li2020deep}, we sample an additional 5,000 labeled pairs from each target dataset to train domain-specific models. The specific datasets are:  Abt-buy (AB), Amazon-Google (AG), DBLP-ACM (DA), DBLP-Scholar (DS), IMDb-TMDb (IM), IMDb-TVDb (IT), TMDb-TVDb (TT), and Walmart-Amazon (WA).

\noindent\textbf{Hardware platforms and their price.}
All experiments involving local computation are conducted on NVIDIA H100 GPUs.
To ensure a realistic and fair economic comparison, we standardize the GPU compute cost at \$5.98 per hour. This rate is derived by averaging the on-demand pricing for H100 instances across four major cloud service providers; more details are provided in Appendix~\ref{appendix:E}.

\noindent\textbf{Evaluation Metrics.}
Traditional pairwise F1 scores compute performance over isolated pairs, ignoring the mutual exclusivity often required in real-world applications, such as an anchor has at most one valid match~\citep{christen2012concepts}. Given that our system makes a holistic decision over the candidate set $(a, C)$, we employ instance level metrics rather than pair level ones. This protocol is strictly more demanding: a ``Success'' is counted only when the exact ground-truth is selected from $K$ candidates; selecting a wrong candidate counts as both a False Positive and a False Negative. Notably, under the assumption of single-match validity, this instance level protocol aligns mathematically with the standard pairwise F1 score. Accordingly, we report $\mathbf{F1_{id}}$ to measure selection accuracy on anchors with valid matches, and $\mathbf{F1_{none}}$ to evaluate rejection sensitivity on those without. We define $\mathbf{F1_{macro}}$ as their unweighted average to ensure a balanced assessment that penalizes both selecting when none exists and misses.

\noindent\textbf{Cost Calculation.}
We perform a comprehensive economic analysis covering both computational and API costs. For LLM-based components, costs are derived from token usage based on the standardized API pricing across multiple platforms, as shown in Table~\ref{tab:models}. For the training of CaRL-EM and the full lifecycle of the supervised baseline \textsc{Ditto}, we incorporate GPU compute costs standardized at \$5.98/hour.

All compared methods use the same candidate sets produced by the same blocking stage. Therefore, blocking does not affect the relative comparison among decision policies; in our setting, its cost is also negligible compared with LLM-based decision cost. For completeness, we report the blocking cost separately in Appendix~\ref{appendix:F}.

During the inference phase, CaRL-EM's policy network is implemented as a lightweight MLP. Its computational overhead on a commodity CPU is negligible compared to the network latency of LLM API calls. Therefore, we assume zero hardware cost for CaRL-EM's inference, as it can be efficiently served without a GPU. For CaRL-EM, we distinguish deployment-time inference cost from one-time offline training cost. The latter is incurred once; we report it transparently in Appendix~\ref{appendix:F}.

\section{Results}

\subsection{Main Results}
\label{sec:main_results}

Table~\ref{tab:main_results} reports instance level $F1_{\text{macro}}$  and cost on 7 datasets. \textsc{Ditto} is trained in-domain, so it is trained separately on each target dataset. In contrast, CaRL-EM is trained only once on the AB dataset and then used zero-shot on the other datasets without any additional fine-tuning. Overall, CaRL-EM keeps $F1_{\text{macro}}$ competitive while using less inference cost, which leads to a higher efficiency score under our metric.

\begin{table*}[t]
\centering
\small
\resizebox{\textwidth}{!}{
\begin{tabular}{lcccccccc|cc|c}
\hline
\textbf{Method} & \textbf{Training} & \textbf{AG} & \textbf{DA} & \textbf{DS} & \textbf{IM} & \textbf{IT} & \textbf{TT} & \textbf{WA}  & \textbf{Avg.} & \textbf{Cost (\$)} & \textbf{Eff. Score} \\
\hline
\multicolumn{12}{l}{\textit{\textbf{Reference: Supervised (In-domain)}}} \\
\textcolor{gray}{\textsc{Ditto}~\citep{li2020deep}} & \textcolor{gray}{per-dataset} & \textcolor{gray}{63.3} & \textcolor{gray}{96.8} & \textcolor{gray}{88.4} & \textcolor{gray}{93.9} & \textcolor{gray}{89.8} & \textcolor{gray}{87.0} & \textcolor{gray}{79.8} & \textcolor{gray}{85.64} & \textcolor{gray}{2.22} & \textcolor{gray}{73.2} \\
\hline
\hline
\multicolumn{12}{l}{\textit{\textbf{Zero-shot / Transfer Settings}}} \\
Matching~\citep{Ralph2025Entity}   & none & 29.18 & 70.88 & 71.27 & 43.91 & 34.15 & 35.44 & 26.09 & 46.66 & 2.19 & 40.2 \\
Comparing  & none & 47.35 & \textbf{85.04} & \textbf{77.89} & 61.66 & 33.60 & 46.29 & 43.71 & 58.08 & 0.54 & 134.4 \\
Selecting  & none & 42.55 & 66.90 & 59.75 & \textbf{92.20} & \textbf{84.00} & 84.25 & 63.70 & 69.94 & \underline{0.15} & \underline{499.7} \\
\textsc{COMEM}~\citep{wang2025match}      & none & \textbf{60.01} & 56.94 & 69.54 & 87.32 & 75.46 & \underline{84.54} & \textbf{78.76} & \underline{74.25} & 0.59 & 160.0 \\
\textbf{CaRL-EM (Ours)} & \textbf{AB-only} & \underline{54.65} & \underline{78.50} & \underline{77.75} & \underline{88.70} & \underline{82.65} & \textbf{85.80} & \underline{64.60} & \textbf{76.09} & \textbf{0.13} & \textbf{623.7} \\
\hline
\end{tabular}
}
\caption{
$F1_{\text{macro}}$ and inference cost on 7 benchmarks.
\textsc{Ditto} is a supervised reference trained in-domain, with a separate model for each dataset.
\textsc{COMEM} is the strongest LLM pipeline baseline in our comparison.
We focus on methods under the zero-shot and transfer settings.
Eff. Score is a log-efficiency metric, defined as $F1_{\text{macro}} / \ln(1+\text{Cost})$, to capture the quality cost trade-off.
}
\label{tab:main_results}
\end{table*}

\noindent\textbf{Comparison with the supervised baseline.}
\textsc{Ditto} is an in-domain supervised model, so it is trained separately on each dataset. CaRL-EM is trained once on AB and then transferred to the other 7 datasets with no extra tuning. Under this setting, CaRL-EM achieves an average $F1_{\text{macro}}$ of 76.09, while \textsc{Ditto} reaches 85.64. This means CaRL-EM scores about 89\% of \textsc{Ditto}. However, the cost of CaRL-EM is 5.9\% of \textsc{Ditto}'s cost. This shows that the CaRL-EM can reuse one policy across datasets and keep the cost low. As we discuss in \S5.3, \textsc{Ditto} does not transfer well across domains without retraining.

\noindent\textbf{Comparison with LLM-based methods.}
Among zero-shot methods, CaRL-EM achieves the best average $F1_{\text{macro}}$. It also outperforms the strongest hand-crafted baseline, \textsc{COMEM}. At the same time, it uses less cost, with about a 4.5$\times$ reduction. Intuitively, a fixed pipeline is more likely to pick a candidate even when the evidence is weak, which leads to false positives. In contrast, CaRL-EM first applies the global \textsc{Select} operator and then uses \textsc{Match} and \textsc{Compare} for local checks. When the information is not enough, it is also more likely to stop with a no-match decision. We further support this point with $F1_{\text{none}}$ in Table~\ref{tab:backends}.

\subsection{Cost-Efficiency Analysis}
\label{sec:cost_analysis}

\begin{figure}[t]
\centering
\includegraphics[width=\columnwidth]{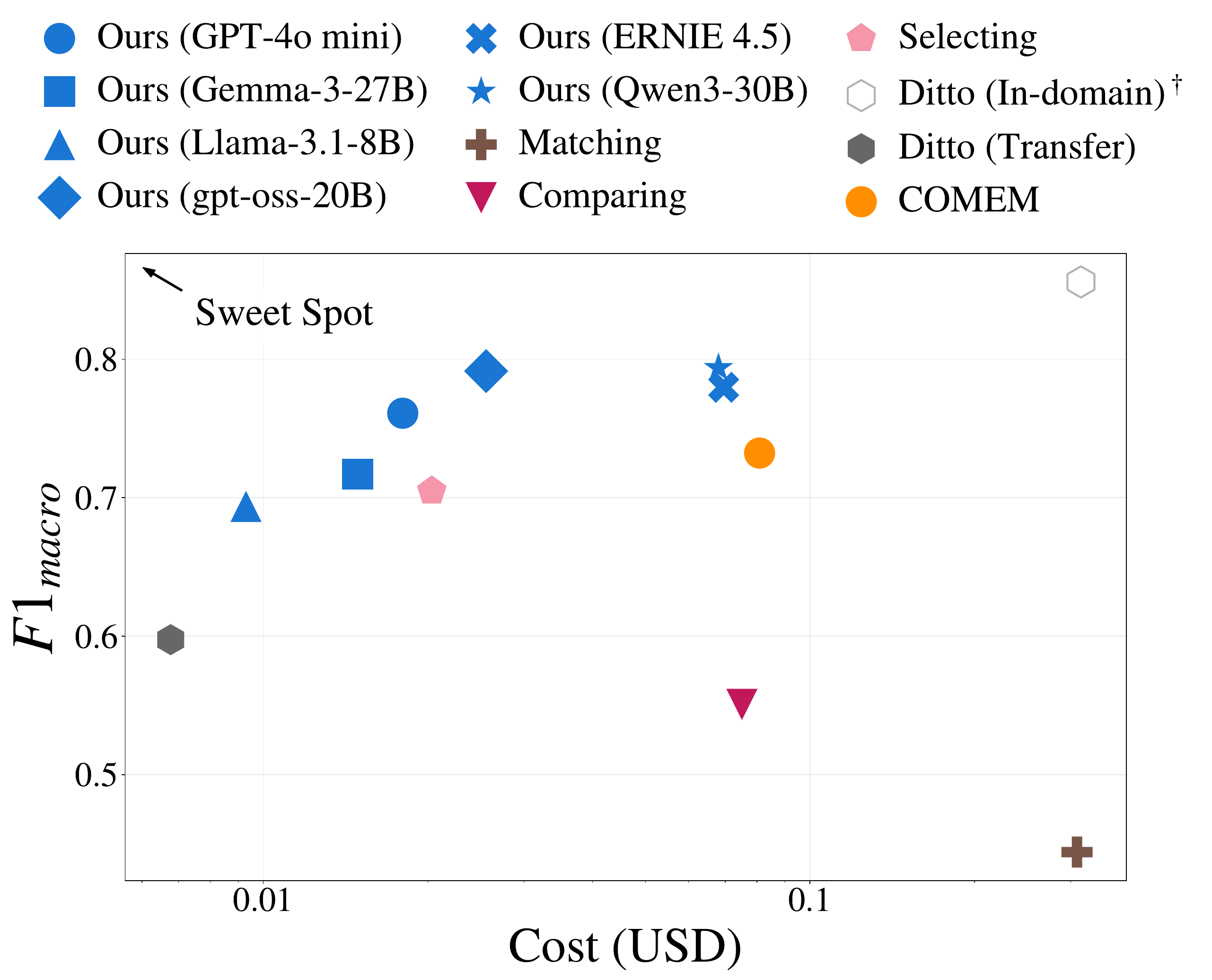}
\caption{
Pareto plot: total cost to process the 7 test datasets and mean $F1_{\text{macro}}$. Points closer to the top-left indicate a better quality cost trade-off. $^\dagger$ \textsc{Ditto} (In-domain) is fine-tuned separately for each dataset; its training and cost accounting also differ from other methods, so we report it only as a reference.
}
\label{fig:pareto}
\end{figure}

\noindent\textbf{Cost quality trade-off.}
Figure~\ref{fig:pareto} compares methods in the dataset level total cost and $F1_{\text{macro}}$.\footnote{We report the cost of inference for all zero-shot methods in this plot. CaRL-EM’s one-time offline cost is reported separately in Appendix~\ref{appendix:F}.} In the figure, CaRL-EM points are mostly in the high-quality, low-cost region. Without retraining the controller, swapping in a stronger backend LLM usually improves $F1_{\text{macro}}$, but it also increases cost. Under the same LLM, \textsc{CaRL-EM (GPT-4o mini)} achieves a higher $F1_{\text{macro}}$ than the state-of-the-art cost-efficient baseline \textsc{COMEM}, while using only 22\% of its average total cost. Overall, CaRL-EM forms a new Pareto frontier~\citep{deb2002nsga2}. This shows that it offers the best quality cost trade-off among the methods we evaluate. The gain mainly comes from the controller’s adaptive decisions, rather than any single LLM.

\subsection{Transfer and Robustness to LLM Backends}
\label{sec:transfer-backends}

\noindent\textbf{Zero-shot transfer.}
Table~\ref{tab:backends} compares CaRL-EM with \textsc{COMEM} and a transfer baseline based on \textsc{Ditto}. \textsc{Ditto} is trained on AB and then applied to the other datasets without retraining, so its $F1_{\text{macro}}$ drops from 85.64 to 59.75. In contrast, CaRL-EM is also not retrained, yet it still achieves a high $F1_{\text{macro}}$. This suggests that the learned decision strategy captures potential patterns in multi-candidate EM that transfer across datasets.

\noindent\textbf{Swapping LLM backends.}
The same controller can also work with different LLM backends. We test this by swapping the backend models at inference time, without retraining the controller. As shown in Table~\ref{tab:backends}, stronger backends usually improve $F1_{\text{macro}}$, but they also increase cost.
The controller mainly adjusts how often it uses \textsc{Match} and \textsc{Compare} to balance quality and cost. This pattern matches our design goal: the policy learns the decision sequence, rather than overfitting to one dataset or one specific LLM.

\begin{table*}[t]
    \centering
    \small
    \resizebox{\textwidth}{!}{
    \begin{tabular}{lccccccl}
    \hline
    \textbf{Backend} &
    \textbf{$F1_{\text{id}}$} &
    \textbf{$F1_{\text{none}}$} &
    \textbf{$F1_{\text{macro}}$} &
    \textbf{Avg.\ $S_{\text{match}}$} &
    \textbf{Avg.\ $S_{\text{compare}}$} &
    \textbf{Avg.\ $S_{\text{select}}$} &
    \textbf{Cost} \\
    \hline
    \textsc{Ditto}$^\dagger$
        & 73.52 & 45.99 & 59.75 &  --   & --   & --   & \textbf{0.05} \\
    \textsc{COMEM}
        & \textbf{85.66} & 60.78 & 73.22 &  10.00 & --  & \underline{1.00}  & 0.59 \\
    CaRL-EM$_L$ (Llama-3.1)
        & 81.23 & 57.49 & 69.36 &  \textbf{1.42}  & 2.03  & \textbf{0.99}  & $\underline{0.07}_{(\sim0.20)}$ \\
    CaRL-EM$_E$ (Gemma3)
        & 82.86 & 60.56 & 71.70 &  \underline{1.69} & 2.02  & \textbf{0.99} & 0.11 \\
    CaRL-EM (GPT-4o mini)$^\ddagger$
        & 83.99 & 68.20 & 76.09 &  1.92  & 1.95  & \textbf{0.99}  & 0.13 \\
    CaRL-EM$_G$ (GPT-oss)
        & \underline{85.41} & \underline{72.91} & \underline{79.14} &  2.03  & 1.91 & \textbf{0.99} & $0.18_{(\sim0.34)}$ \\
    CaRL-EM$_E$ (ERNIE 4.5)
        & 83.77 & 72.17 & 77.97 &  2.47  & \textbf{1.89}  & \textbf{0.99}  & 0.49 \\
    CaRL-EM$_Q$ (Qwen3)
        & \underline{85.41} & \textbf{73.40} & \textbf{79.37} &  2.18  & \underline{1.90}  & \textbf{0.99}  & $0.48_{(\sim1.57)}$ \\
    \hline
    \end{tabular}
    }
    \caption{
    Transfer and backend robustness on the seven target datasets. We report $F1_{\text{id}}$, $F1_{\text{none}}$, and $F1_{\text{macro}}$, as well as the average number of operator calls per episode ($S_{\text{match}}$, $S_{\text{compare}}$, $S_{\text{select}}$). $^\dagger$ \textsc{Ditto} is trained on AB and applied to the other datasets without retraining. $^\ddagger$ CaRL-EM is trained once on AB with GPT-4o mini, and we swap the backend LLMs at test time without retraining the controller.
    }
    \label{tab:backends}
\end{table*}

\subsection{Controller Behavior and Position Bias}
\label{sec:behavior}

We inspect the learned policy to understand how it manages LLM operators and how this affects robustness to position bias in long candidate lists~\citep{liu2024lost}.

\begin{figure}[t]
    \centering
    \includegraphics[width=\columnwidth]{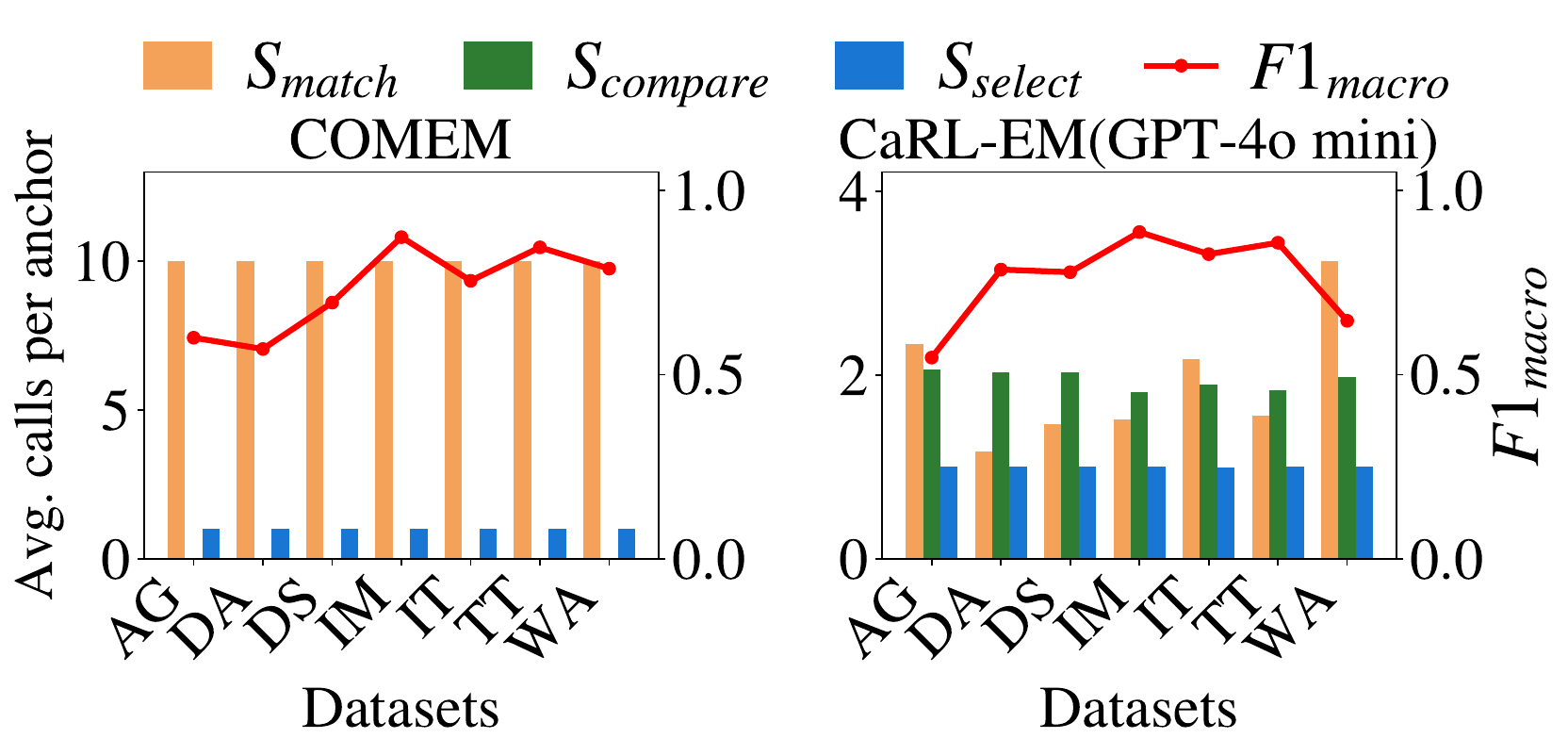}
    \caption{
    Average number of operator calls per anchor on each dataset.
    }
    \label{fig:operator-usage}
\end{figure}

\noindent\textbf{Operator usage patterns.}
Figure~\ref{fig:operator-usage} shows the average number of operator calls per anchor. Compared with \textsc{COMEM}, CaRL-EM uses fewer \textsc{Match} and \textsc{Compare} calls. At the same time, the amount of tool usage varies across domains and datasets, showing that the learned policy does not follow a fixed call pattern. We also observe that on relatively harder datasets such as AG and WA, CaRL-EM tends to run more steps before making a final decision. In contrast, \textsc{COMEM} follows a fixed pipeline and cannot adjust its strategy across instances or datasets.

\begin{figure}[t]
    \centering
    \includegraphics[width=\columnwidth]{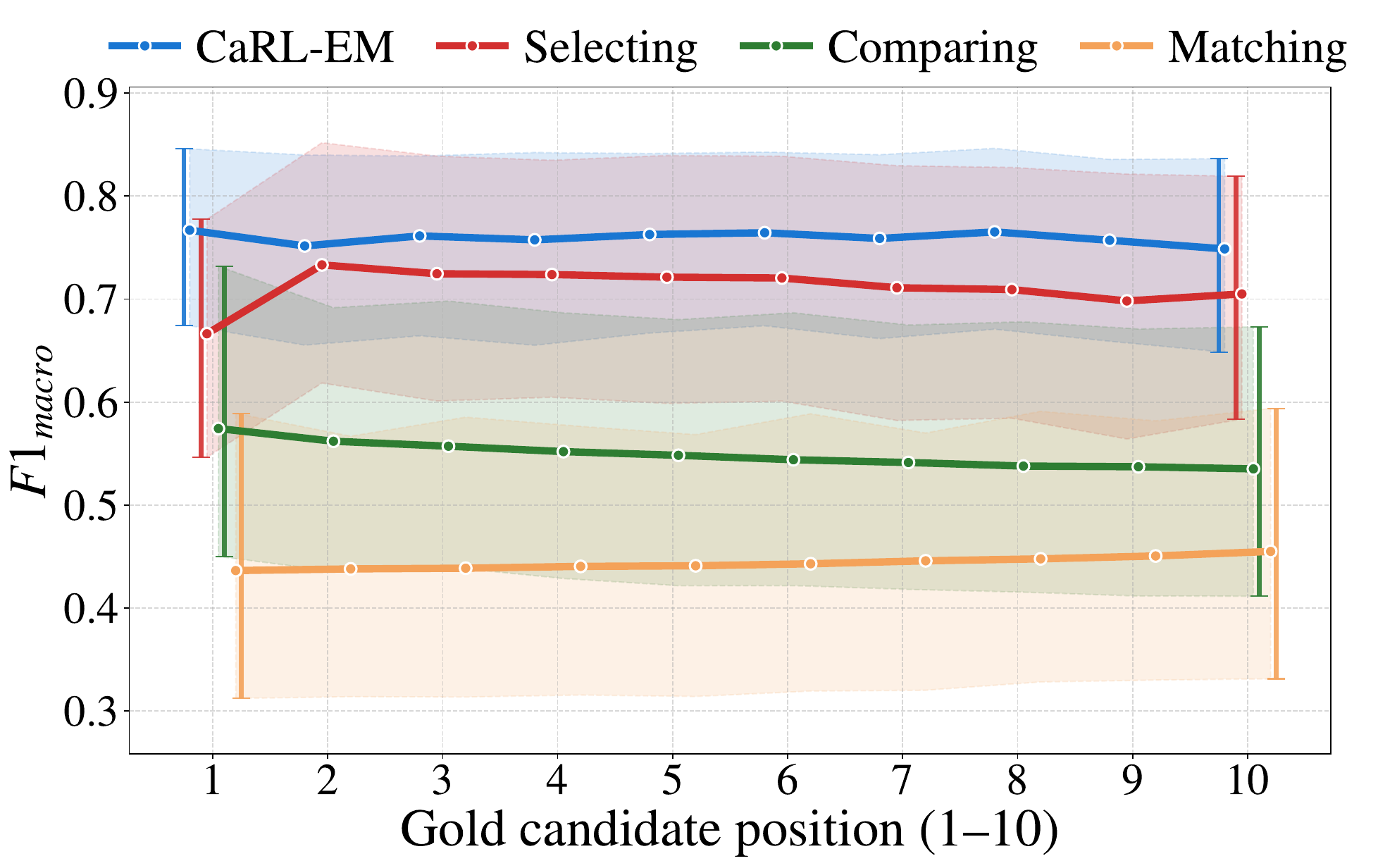}
    \caption{
    $F1_{\text{macro}}$ as a function of the gold candidate's initial position in the list. Lines show the mean performance at each position. Shaded regions indicate variation across datasets.
    }
    \label{fig:position-bias}
\end{figure}

\begin{figure*}[t]
  \includegraphics[width=\textwidth]{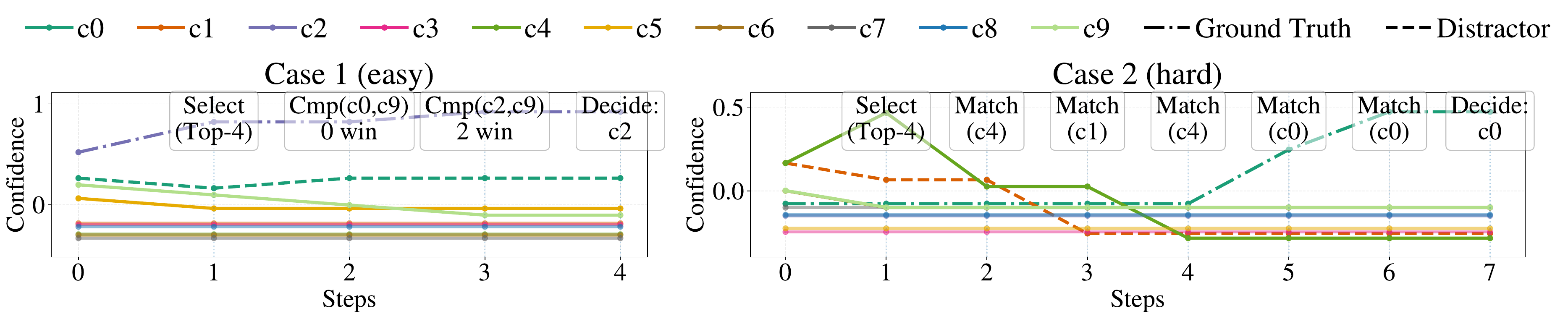}
  \caption{
  Two cases from AG processed by CaRL-EM. Colors denote different candidates. Lines show how each candidate's confidence changes with each action. Initial confidence comes from a similarity score between the anchor and the candidates.
}
  \label{fig:case}
\end{figure*}

\noindent\textbf{Mitigating long-context position bias.}
We run a controlled test to isolate position effects in the 7 test datasets. For each anchor, we force the gold candidate to appear at a fixed position (0--9) in the initial list. We keep the gold position fixed and shuffle the other candidates. We repeat this process 10 times for each position and report the average result.

As shown in Figure~\ref{fig:position-bias}, the Selecting baseline is more sensitive to where the gold candidate appears in the list. It performs notably worse when the gold candidate is placed at position 1, and it still varies across other positions. The Matching and Matching baselines are flatter, but Comparing declines slightly, while Matching increases slightly. The shaded bands indicate gaps across datasets. This gap is larger for Matching and Comparing, which suggests weaker stability across domains. In contrast, CaRL-EM stays more stable across positions and shows the smallest cross-dataset variation. A key reason is that CaRL-EM does not rely on a single long listwise call. It can flexibly combine \textsc{Match}/\textsc{Compare} and \textsc{Select}, and it can perform mutual verification between various operators, which reduces dependence on the initial order.

\subsection{Qualitative Analysis: Policy Behavior}
\label{sec:qualitative}

To understand how CaRL-EM achieves efficiency, we visualize the decision trajectory on two cases from the AG dataset, as shown in Figure~\ref{fig:case}.

\noindent\textbf{Case 1 (easy):} From the initial confidence scores, the gold candidate (c2) has a clear advantage. The policy first calls \textsc{Select} on the top-4 list. It then uses two \textsc{Compare} calls as quick checks.
It finally decides on c2. This takes only a few steps and avoids checking all candidates one by one; in this case, the policy does not call the pairwise \textsc{Match} operator.

\noindent\textbf{Case 2 (Hard):} In this case, several candidates start with similar scores, and the gold candidate (c0) doesn't have the highest confidence score. The policy again calls \textsc{Select} on the top-4 list, but the first result is c4. It then calls \textsc{Match} for local verification and finds that it conflicts with the result from the previous step. Therefore, it runs multiple \textsc{Match} calls and gradually improves the confidence of c0. This shows that the policy spends more local checks on hard cases, but stops early on easy ones, which allocates the budget based on instance difficulty. This adaptive behavior demonstrates that CaRL-EM acts as a "System 2" thinker, allocating computational resources dynamically based on instance difficulty~\citep{kahneman2011thinking}.

\subsection{Ablation and Design Choices}
\label{sec:ablation}

We run ablations to understand which components matter. We retrain each variant on AB and report results averaged over the other seven datasets. Table~\ref{tab:ablation} summarizes accuracy, stability (Std.), and normalized inference cost.

\noindent\textbf{Cost-aware reward.}
Removing the cost term keeps $F1_{\text{macro}}$ almost unchanged, but the cost increases by 15\%. This shows that the cost term is needed to learn a more efficient policy.

\noindent\textbf{Potential-based shaping.}
Removing the global potential term $\phiGlob_t$ reduces the $F1_{\text{id}}$ and $F1_{\text{macro}}$ value and makes performance unstable. This shows that the shaping signal helps the policy separate true and false candidates during the episode.

\noindent\textbf{Operators.}
Each operator plays a different role. Without \textsc{Match}, the model is cheaper but its accuracy drops, especially on none cases. Without \textsc{Select}, the model is the cheapest and $F1_{\text{none}}$ is high, but $F1_{\text{id}}$ drops and overall performance is lower. Without \textsc{Compare}, performance also drops, which suggests that pairwise checks help when top candidates are close.

\begin{table}[t]
    \centering
    \setlength{\tabcolsep}{4pt}
    \small
    \begin{tabular}{lccccc}
    \hline
    \textbf{Variant} &
    $F1_{\text{id}}$ &
    $F1_{\text{none}}$ &
    $F1_{\text{macro}}$ &
    Std. &
    Cost \\
    \hline
    Full (ours)
        & \textbf{83.99} & 68.20 & \textbf{76.09} &  \textbf{12.24} & 0.13 \\
    w/o cost
        & 82.96 & \underline{69.16} & \underline{76.07} & 13.05 & 0.15 \\
    w/o $\phiGlob_t$
        & \underline{83.03} & 68.89 & 75.93  & 14.56  & 0.13 \\
    w/o Match
        & 81.91 & 49.86 & 65.87 &  \underline{12.59} & \underline{0.10} \\
    w/o Compare
        & 80.04 & 67.31 & 73.69 & 15.04 & 0.13 \\
    w/o Select
        & 75.13 & \textbf{71.36} & 73.26 & 20.60 & \textbf{0.08}\\
    \hline
    \end{tabular}
    \caption{
      Ablation study of CaRL-EM variants. We report instance level $F1_{\text{id}}$, $F1_{\text{none}}$, macro-average $F1_{\text{macro}}$, the standard deviation across benchmarks, and inference cost, all averaged over the 7 benchmarks.
    }
    \label{tab:ablation}
\end{table}

\section{Conclusion}
We study LLM-based EM in the practical blocked setting, where each anchor comes with a small candidate set and cost quickly becomes the bottleneck at scale. We propose CaRL-EM, a reinforcement learning controller that treats EM as a sequential decision process and decides when to use \textsc{Match}, \textsc{Compare}, \textsc{Select}, or \textsc{Decide}, while also choosing between cheaper and stronger model capacities. To our knowledge, this is the first work that casts LLM-driven EM as a cost-aware sequential RL problem.

Across seven benchmarks under zero-shot transfer, CaRL-EM learns to spend less on easy cases and check more on hard ones, and it yields a better quality cost trade-off than strong LLM baselines and hand-designed pipelines. More broadly, our formulation offers a way to think about trading off decision strategy performance and cost through a learned controller, rather than a fixed pipeline.

\section*{Limitations}

Our study focuses on the clean-clean setting where an anchor has at most one true match. While this assumption fits many standard benchmarks and blocking-based pipelines, it does not cover settings with multiple valid matches, one-to-many links, or noisy/duplicate-heavy tables. Extending CaRL-EM to those cases would likely require changes to both the state (e.g., tracking multiple plausible matches) and the stopping/decision rule.

In addition, we mainly consider small candidate pools (top-10 in our protocol). When candidate sets become much larger, a controller that directly reasons over the whole list may become less effective, and a different design (e.g., multi-stage pruning, hierarchical control, or tighter coupling with retrieval) may be needed to keep both cost and decision quality under control.

Finally, CaRL-EM is trained with a coarse two level abstract cost model. This makes backend swapping simple and keeps the policy less tied to a specific pricing scheme, but it does not capture the full complexity of real deployment costs, such as prompt-length differences, token-based billing, latency constraints, batching effects, and provider-specific pricing. As a result, the learned behavior may not be cost-optimal under a different cost surface, and deployments may require recalibrating the cost function (or retraining) to match the target environment.

\section*{Ethical Considerations}
We do not collect new data. Our experiments use public entity-matching benchmarks and we do not release any additional data.
The main risk is incorrect matches, which can lead to wrong merges and downstream errors. High-stakes use should include auditing and human review. We discourage using entity matching to link personal identities across datasets and emphasize legal and ethical compliance.

\section*{Acknowledgments}
We thank Frank van Harmelen for his valuable feedback and suggestions on this paper.
We also thank the anonymous reviewers for their constructive comments.
Chaohui Guo is supported by the China Scholarship Council (CSC).

\bibliography{custom}

@inproceedings{peeters2023using,
  title={Using chatgpt for entity matching},
  author={Peeters, Ralph and Bizer, Christian},
  booktitle={European Conference on Advances in Databases and Information Systems},
  pages={221--230},
  year={2023},
  organization={Springer}
}

@article{li2024leveraging,
  title={On leveraging large language models for enhancing entity resolution: a cost-efficient approach},
  author={Li, Huahang and Feng, Longyu and Li, Shuangyin and Hao, Fei and Zhang, Chen Jason and Song, Yuanfeng},
  journal={arXiv preprint arXiv:2401.03426},
  year={2024}
}

@misc{christen2012concepts,
  title={Data matching: concepts and techniques for record linkage, entity resolution, and duplicate detection.},
  author={Peter Christen},
    year={2012},
  publisher={Springer: Data-centric systems and applications}
}

@article{ebraheem2017deeper,
  title={DeepER--Deep Entity Resolution},
  author={Ebraheem, Muhammad and Thirumuruganathan, Saravanan and Joty, Shafiq and Ouzzani, Mourad and Tang, Nan},
  journal={arXiv preprint arXiv:1710.00597},
  year={2017}
}

@inproceedings{mudgal2018deep,
  title={Deep learning for entity matching: A design space exploration},
  author={Mudgal, Sidharth and Li, Han and Rekatsinas, Theodoros and Doan, AnHai and Park, Youngchoon and Krishnan, Ganesh and Deep, Rohit and Arcaute, Esteban and Raghavendra, Vijay},
  booktitle={Proceedings of the 2018 international conference on management of data},
  pages={19--34},
  year={2018}
}

@article{li2020deep,
  title={Deep entity matching with pre-trained language models},
  author={Li, Yuliang and Li, Jinfeng and Suhara, Yoshihiko and Doan, AnHai and Tan, Wang-Chiew},
  journal={arXiv preprint arXiv:2004.00584},
  year={2020}
}

@inproceedings{wang2025match,
  title={Match, compare, or select? an investigation of large language models for entity matching},
  author={Wang, Tianshu and Chen, Xiaoyang and Lin, Hongyu and Chen, Xuanang and Han, Xianpei and Sun, Le and Wang, Hao and Zeng, Zhenyu},
  booktitle={Proceedings of the 31st International Conference on Computational Linguistics},
  pages={96--109},
  year={2025}
}

@article{feng2025retool,
  title={Retool: Reinforcement learning for strategic tool use in llms},
  author={Feng, Jiazhan and Huang, Shijue and Qu, Xingwei and Zhang, Ge and Qin, Yujia and Zhong, Baoquan and Jiang, Chengquan and Chi, Jinxin and Zhong, Wanjun},
  journal={arXiv preprint arXiv:2504.11536},
  year={2025}
}

@article{zhang2024treacle,
  title={TREACLE: Thrifty Reasoning via Context-Aware LLM and Prompt Selection},
  author={Zhang, Xuechen and Huang, Zijian and Taga, Ege Onur and Joe-Wong, Carlee and Oymak, Samet and Chen, Jiasi},
  journal={CoRR},
  year={2024}
}

@article{shahbazi2023fairness,
author = {Shahbazi, Nima and Danevski, Nikola and Nargesian, Fatemeh and Asudeh, Abolfazl and Srivastava, Divesh},
title = {Through the Fairness Lens: Experimental Analysis and Evaluation of Entity Matching},
year = {2023},
issue_date = {July 2023},
publisher = {VLDB Endowment},
volume = {16},
number = {11},
issn = {2150-8097},
url = {https://doi.org/10.14778/3611479.3611525},
doi = {10.14778/3611479.3611525},
journal = {Proc. VLDB Endow.},
month = jul,
pages = {3279–3292},
numpages = {14}
}

@article{papadakis2020blocking,
author = {Papadakis, George and Skoutas, Dimitrios and Thanos, Emmanouil and Palpanas, Themis},
title = {Blocking and Filtering Techniques for Entity Resolution: A Survey},
year = {2020},
issue_date = {March 2021},
publisher = {Association for Computing Machinery},
address = {New York, NY, USA},
volume = {53},
number = {2},
issn = {0360-0300},
url = {https://doi.org/10.1145/3377455},
doi = {10.1145/3377455},
journal = {ACM Comput. Surv.},
month = mar,
articleno = {31},
numpages = {42}
}

@article{thirumuruganathan2021deepblocking,
author = {Thirumuruganathan, Saravanan and Li, Han and Tang, Nan and Ouzzani, Mourad and Govind, Yash and Paulsen, Derek and Fung, Glenn and Doan, AnHai},
title = {Deep learning for blocking in entity matching: a design space exploration},
year = {2021},
issue_date = {July 2021},
publisher = {VLDB Endowment},
volume = {14},
number = {11},
issn = {2150-8097},
url = {https://doi.org/10.14778/3476249.3476294},
doi = {10.14778/3476249.3476294},
journal = {Proc. VLDB Endow.},
month = jul,
pages = {2459–2472},
numpages = {14}
}

@article{paulsen2023sparkly,
author = {Paulsen, Derek and Govind, Yash and Doan, AnHai},
title = {Sparkly: A Simple yet Surprisingly Strong TF/IDF Blocker for Entity Matching},
year = {2023},
issue_date = {February 2023},
publisher = {VLDB Endowment},
volume = {16},
number = {6},
issn = {2150-8097},
url = {https://doi.org/10.14778/3583140.3583163},
doi = {10.14778/3583140.3583163},
journal = {Proc. VLDB Endow.},
month = feb,
pages = {1507–1519},
numpages = {13}
}

@article{barlaug2020survey,
author = {Barlaug, Nils and Gulla, Jon Atle},
title = {Neural Networks for Entity Matching: A Survey},
year = {2021},
issue_date = {June 2021},
publisher = {Association for Computing Machinery},
address = {New York, NY, USA},
volume = {15},
number = {3},
issn = {1556-4681},
url = {https://doi.org/10.1145/3442200},
doi = {10.1145/3442200},
journal = {ACM Trans. Knowl. Discov. Data},
month = apr,
articleno = {52},
numpages = {37}
}

@inproceedings{shah2018extreme,
    title = "Neural Network based Extreme Classification and Similarity Models for Product Matching",
    author = "Shah, Kashif  and
      Kopru, Selcuk  and
      Ruvini, Jean-David",
    editor = "Bangalore, Srinivas  and
      Chu-Carroll, Jennifer  and
      Li, Yunyao",
    booktitle = "Proceedings of the 2018 Conference of the North {A}merican Chapter of the Association for Computational Linguistics: Human Language Technologies, Volume 3 (Industry Papers)",
    month = jun,
    year = "2018",
    address = "New Orleans - Louisiana",
    publisher = "Association for Computational Linguistics",
    url = "https://aclanthology.org/N18-3002/",
    doi = "10.18653/v1/N18-3002",
    pages = "8--15"
}

@inproceedings{tracz2020bert,
    title = "{BERT}-based similarity learning for product matching",
    author = "Tracz, Janusz  and
      W{\'o}jcik, Piotr Iwo  and
      Jasinska-Kobus, Kalina  and
      Belluzzo, Riccardo  and
      Mroczkowski, Robert  and
      Gawlik, Ireneusz",
    editor = "Zhao, Huasha  and
      Sondhi, Parikshit  and
      Bach, Nguyen  and
      Hewavitharana, Sanjika  and
      He, Yifan  and
      Si, Luo  and
      Ji, Heng",
    booktitle = "Proceedings of Workshop on Natural Language Processing in E-Commerce",
    month = dec,
    year = "2020",
    address = "Barcelona, Spain",
    publisher = "Association for Computational Linguistics",
    url = "https://aclanthology.org/2020.ecomnlp-1.7/",
    pages = "66--75"
}

@article{peeters2021dualobjective,
author = {Peeters, Ralph and Bizer, Christian},
title = {Dual-objective fine-tuning of BERT for entity matching},
year = {2021},
issue_date = {June 2021},
publisher = {VLDB Endowment},
volume = {14},
number = {10},
issn = {2150-8097},
url = {https://doi.org/10.14778/3467861.3467878},
doi = {10.14778/3467861.3467878},
journal = {Proc. VLDB Endow.},
month = jun,
pages = {1913–1921},
numpages = {9}
}

@book{kahneman2011thinking,
  author    = {Kahneman, Daniel},
  title     = {Thinking, Fast and Slow},
  year      = {2011},
  publisher = {Farrar, Straus and Giroux},
  address   = {New York},
  isbn      = {9780374275631}
}

@article{Konda2016Magellan,
author = {Konda, Pradap and Das, Sanjib and Suganthan G. C., Paul and Doan, AnHai and Ardalan, Adel and Ballard, Jeffrey R. and Li, Han and Panahi, Fatemah and Zhang, Haojun and Naughton, Jeff and Prasad, Shishir and Krishnan, Ganesh and Deep, Rohit and Raghavendra, Vijay},
title = {Magellan: toward building entity matching management systems},
year = {2016},
issue_date = {August 2016},
publisher = {VLDB Endowment},
volume = {9},
number = {12},
issn = {2150-8097},
url = {https://doi.org/10.14778/2994509.2994535},
doi = {10.14778/2994509.2994535},
journal = {Proc. VLDB Endow.},
month = aug,
pages = {1197–1208},
numpages = {12}
}

@article{chen2023frugalgpt,
  title={Frugalgpt: How to use large language models while reducing cost and improving performance},
  author={Chen, Lingjiao and Zaharia, Matei and Zou, James},
  journal={arXiv preprint arXiv:2305.05176},
  year={2023}
}

@article{ong2025routellm,
  title={Routellm: Learning to route llms with preference data, 2024},
  author={Ong, Isaac and Almahairi, Amjad and Wu, Vincent and Chiang, Wei-Lin and Wu, Tianhao and Gonzalez, Joseph E and Kadous, M Waleed and Stoica, Ion},
  journal={URL https://arxiv. org/abs/2406.18665},
  volume={4},
  year={2025}
}

@article{Raffel2020Exploring,
author = {Raffel, Colin and Shazeer, Noam and Roberts, Adam and Lee, Katherine and Narang, Sharan and Matena, Michael and Zhou, Yanqi and Li, Wei and Liu, Peter J.},
title = {Exploring the limits of transfer learning with a unified text-to-text transformer},
year = {2020},
issue_date = {January 2020},
publisher = {JMLR.org},
volume = {21},
number = {1},
issn = {1532-4435},
journal = {J. Mach. Learn. Res.},
month = jan,
articleno = {140},
numpages = {67}
}

@book{sutton1998reinforcement,
  title={Reinforcement learning: An introduction},
  author={Sutton, Richard S and Barto, Andrew G and others},
  volume={1},
  number={1},
  year={1998},
  publisher={MIT press Cambridge}
}

@article{schulman2017proximal,
  title={Proximal policy optimization algorithms},
  author={Schulman, John and Wolski, Filip and Dhariwal, Prafulla and Radford, Alec and Klimov, Oleg},
  journal={arXiv preprint arXiv:1707.06347},
  year={2017}
}

@article{gemmell2011improving,
  title={Improving entity resolution with global constraints},
  author={Gemmell, Jim and Rubinstein, Benjamin IP and Chandra, Ashok K},
  journal={arXiv preprint arXiv:1108.6016},
  year={2011}
}

@inproceedings{Ralph2025Entity,
  author       = {Ralph Peeters and
                  Aaron Steiner and
                  Christian Bizer},
  editor       = {Alkis Simitsis and
                  Bettina Kemme and
                  Anna Queralt and
                  Oscar Romero and
                  Petar Jovanovic},
  title        = {Entity Matching using Large Language Models},
  booktitle    = {Proceedings 28th International Conference on Extending Database Technology,
                  {EDBT} 2025, Barcelona, Spain, March 25-28, 2025},
  pages        = {529--541},
  publisher    = {OpenProceedings.org},
  year         = {2025},
  url          = {https://doi.org/10.48786/edbt.2025.42},
  doi          = {10.48786/EDBT.2025.42},
  bibsource    = {dblp computer science bibliography, https://dblp.org}
}

@article{grattafiori2024llama,
  title={The llama 3 herd of models},
  author={Grattafiori, Aaron and Dubey, Abhimanyu and Jauhri, Abhinav and Pandey, Abhinav and Kadian, Abhishek and Al-Dahle, Ahmad and Letman, Aiesha and Mathur, Akhil and Schelten, Alan and Vaughan, Alex and others},
  journal={arXiv preprint arXiv:2407.21783},
  year={2024}
}

@article{liu2024lost,
    title = "Lost in the Middle: How Language Models Use Long Contexts",
    author = "Liu, Nelson F.  and
      Lin, Kevin  and
      Hewitt, John  and
      Paranjape, Ashwin  and
      Bevilacqua, Michele  and
      Petroni, Fabio  and
      Liang, Percy",
    journal = "Transactions of the Association for Computational Linguistics",
    volume = "12",
    year = "2024",
    address = "Cambridge, MA",
    publisher = "MIT Press",
    url = "https://aclanthology.org/2024.tacl-1.9/",
    doi = "10.1162/tacl_a_00638",
    pages = "157--173"
}

@inproceedings{ng1999policy,
author = {Ng, Andrew Y. and Harada, Daishi and Russell, Stuart J.},
title = {Policy Invariance Under Reward Transformations: Theory and Application to Reward Shaping},
year = {1999},
isbn = {1558606122},
publisher = {Morgan Kaufmann Publishers Inc.},
address = {San Francisco, CA, USA},
booktitle = {Proceedings of the Sixteenth International Conference on Machine Learning},
pages = {278–287},
numpages = {10},
series = {ICML '99}
}

@article{wiewiora2003potential,
author = {Wiewiora, Eric},
title = {Potential-based shaping and Q-value initialization are equivalent},
year = {2003},
issue_date = {July 2003},
publisher = {AI Access Foundation},
address = {El Segundo, CA, USA},
volume = {19},
number = {1},
issn = {1076-9757},
journal = {J. Artif. Int. Res.},
month = sep,
pages = {205–208},
numpages = {4}
}

@ARTICLE{deb2002nsga2,
  author={Deb, K. and Pratap, A. and Agarwal, S. and Meyarivan, T.},
  journal={IEEE Transactions on Evolutionary Computation}, 
  title={A fast and elitist multiobjective genetic algorithm: NSGA-II}, 
  year={2002},
  volume={6},
  number={2},
  pages={182-197},
  doi={10.1109/4235.996017}}

@article{team2025gemma,
  title={Gemma 3 technical report},
  author={Team, Gemma and Kamath, Aishwarya and Ferret, Johan and Pathak, Shreya and Vieillard, Nino and Merhej, Ramona and Perrin, Sarah and Matejovicova, Tatiana and Ram{\'e}, Alexandre and Rivi{\`e}re, Morgane and others},
  journal={arXiv preprint arXiv:2503.19786},
  year={2025}
}

@article{yang2025qwen3,
  title={Qwen3 technical report},
  author={Yang, An and Li, Anfeng and Yang, Baosong and Zhang, Beichen and Hui, Binyuan and Zheng, Bo and Yu, Bowen and Gao, Chang and Huang, Chengen and Lv, Chenxu and others},
  journal={arXiv preprint arXiv:2505.09388},
  year={2025}
}

@misc{ernie2025technicalreport,
  title        = {ERNIE 4.5 Technical Report},
  author       = {ERNIE Team, Baidu},
  year         = {2025},
  primaryClass = {cs.CL},
  howpublished = {\url{https://yiyan.baidu.com/blog/publication/ERNIE_Technical_Report.pdf}}
}

@article{agarwal2025gpt,
  title={gpt-oss-120b \& gpt-oss-20b model card},
  author={Agarwal, Sandhini and Ahmad, Lama and Ai, Jason and Altman, Sam and Applebaum, Andy and Arbus, Edwin and Arora, Rahul K and Bai, Yu and Baker, Bowen and Bao, Haiming and others},
  journal={arXiv preprint arXiv:2508.10925},
  year={2025}
}

@article{chung2024scaling,
  title={Scaling instruction-finetuned language models},
  author={Chung, Hyung Won and Hou, Le and Longpre, Shayne and Zoph, Barret and Tay, Yi and Fedus, William and Li, Yunxuan and Wang, Xuezhi and Dehghani, Mostafa and Brahma, Siddhartha and others},
  journal={Journal of Machine Learning Research},
  volume={25},
  number={70},
  pages={1--53},
  year={2024}
}

@misc{team2024gpt,
  title={GPT-4o mini: advancing costefficient intelligence},
  author={OpenAI},
  year={2024},
  primaryClass = {cs.CL},
  howpublished = {\url{https://openai.com/index/gpt-4o-mini-advancing-cost-efficient-intelligence/}}
}

\appendix
\section{Hyperparameter Values}
\label{appendix:A}
Table~\ref{tab:hyperparams_carl_em} lists the concrete hyperparameter values used in our experiments.

\begin{table}[t]
\centering
\small
\begin{tabular}{l l l}
\hline
\textbf{Symbol} & \textbf{Meaning} & \textbf{Value} \\
\hline
$N$ & max \#candidates & $10$ \\
$H$ & action history length & $10$ \\
$T_{\max}$ & max decision steps per anchor & $18$ \\
$\rho$ & cost ratio $\kappa_h/\kappa_\ell$ & $2.5$ \\
$k$ & top-$k$ shortlist size(s) for \textsc{Select} & $\{4\}$ \\
$\lambda$ & per-step cost penalty weight in reward & $1.0$ \\
$\eta$ & global potential shaping weight & $0.3$ \\
$\eta_k$ & local shaping weight for \textsc{Select} & $0.5\eta$ \\
$\gamma$ & shaping discount factor & $0.99$ \\
$\sigma$ & Gaussian noise & $0.005$ \\
\hline
\end{tabular}
\caption{Hyperparameter values for CaRL-EM .}
\label{tab:hyperparams_carl_em}
\end{table}

\begin{table*}[t]
    \centering
    \small
    \resizebox{\textwidth}{!}{
    \begin{tabular}{lcccccccl}
    \hline
    \textbf{$\rho$} &
    \textbf{$F1_{\text{id}}$} &
    \textbf{$F1_{\text{none}}$} &
    \textbf{$F1_{\text{macro}}$} &
    \textbf{Avg.\ $S_{\text{match}}$} &
    \textbf{Avg.\ $S_{\text{compare}}$} &
    \textbf{Avg.\ $S_{\text{select}}$} &
    \textbf{Std.} &
    \textbf{Cost} \\
    \hline
    2.5
        & 83.99 & 68.20 & 76.09 &  1.92  & 1.95  & \textbf{0.99}  & 12 & 0.13 \\
    5
        & 83.17 & 71.23 & 77.20 & 2.42  & 1.58  & 0.99 & 15 & \textbf{0.13} \\
    \hline
    \end{tabular}
}
\caption{Sensitivity to the abstract cost ratio $\rho$ (mean over the seven target benchmarks, zero-shot transfer).}
\label{tab:rho_sens}
\end{table*}

\begin{table*}[t]
\centering
\small
\begin{tabular}{lccccc}
\hline
\textbf{Method} & \textbf{Blocking (\$)} & \textbf{Training (\$)} & \textbf{Test (\$)} & \textbf{Total E2E (\$)} & \textbf{Note} \\
\hline
\textsc{Ditto} & 0.00503 & 2.175258222 & 0.04777355556 & 2.228061778 & per-target training \\
Matching & 0.00503 & 0 & 2.19 & 2.19503 & - \\
Comparing & 0.00503 & 0 & 0.54 & 0.54503 & - \\
Selecting & 0.00503 & 0 & 0.15 & 0.15503 & - \\
\textsc{COMEM} & 0.00503 & 0 & 0.59 & 0.59503 & - \\
CaRL-EM & 0.00503 & 16.45 & 0.13 & 16.58503 & train once on AB, reused across targets \\
\hline
\end{tabular}
\caption{End-to-end cost breakdown on the 7 datasets. End-to-end cost includes blocking, training, and test. All methods use the same blocked candidate sets, so blocking cost is shared and negligible in our setting.}
\label{tab:e2e_cost}
\end{table*}

\begin{figure*}[!t]
  \centering
  \includegraphics[width=0.9\textwidth]{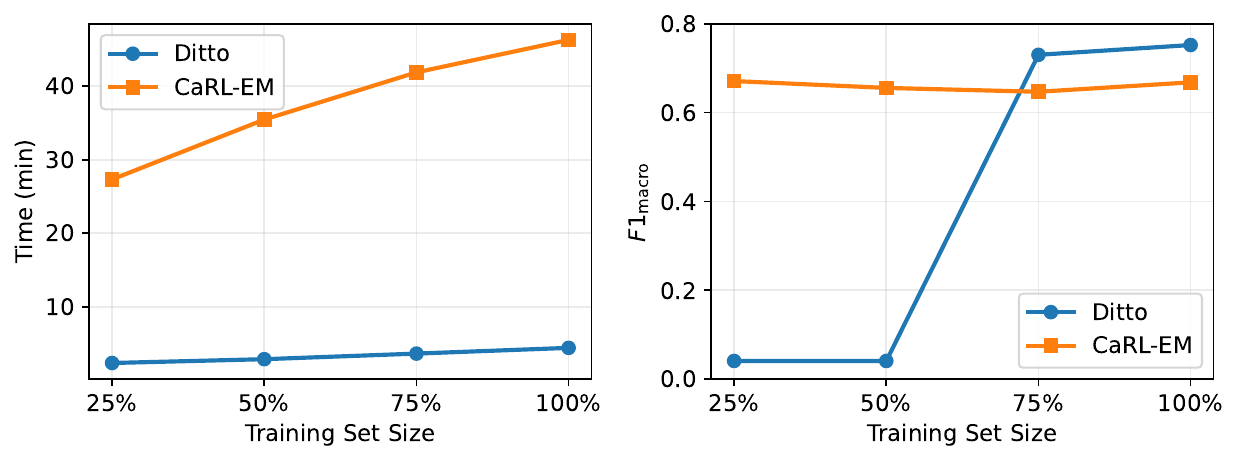}
  \caption{Training scalability on WA under different fine-tuning set sizes. Left: wall-clock fine-tuning time. Right: downstream $F1_{\text{macro}}$ on the WA test set. CaRL-EM requires more wall-clock time because rollout collection repeatedly invokes backend operators, while its performance remains relatively stable across training sizes. \textsc{Ditto} is faster to fine-tune, but under this protocol its performance drops sharply at 25\% and 50\% training sizes.}
  \label{fig:train_scaling_combined}
\end{figure*}

\section{Prompt Templates}
\label{appendix:B}

\subsection{Match}
\noindent\textbf{Output constraint:} the first line must be exactly one token from \texttt{[YES]} or \texttt{[NO]}.

\begin{Verbatim}[fontsize=\small,breaklines=true]
Decide if the two records refer to the SAME real-world entity.
Output EXACTLY ONE token on the first line: [YES] for same, [NO] for different.
[ANCHOR]
{anchor}
[CANDIDATE]
{candidate}
Answer:
\end{Verbatim}

\subsection{Compare}
\noindent\textbf{Output constraint:} the first line must be exactly one token from \texttt{[0]} or \texttt{[1]}.

\begin{Verbatim}[fontsize=\small,breaklines=true]
Which candidate better matches the anchor?
Return EXACTLY ONE token: [0] for the first candidate, [1] for the second.
[ANCHOR]
{anchor}
[0]
{a}
[1]
{b}
Answer:
\end{Verbatim}

\subsection{Select}
\noindent\textbf{Output constraint:} the first line must be exactly one token from \texttt{[0..n-1]} or \texttt{[NONE]}.

\begin{Verbatim}[fontsize=\small,breaklines=true]
Return only one of [0..n-1] or [NONE].
Select the single best-matching candidate for the anchor.
Answer with EXACTLY one token on the first line: [0..{n-1}] or [NONE].
Do NOT include any other words or punctuation.
[ANCHOR]
{anchor}
OPTIONS:
[0] {c0}
[1] {c1}
...
[n-1] {cn-1}
\end{Verbatim}

\section{Sensitivity to the Abstract Cost Ratio $\rho$}
\label{appendix:C}

We test the robustness of the two-level abstract cost ratio $\rho$ by changing it from $\rho{=}2.5$ to $\rho{=}5$, retraining CaRL-EM on Abt-Buy (AB) and evaluating zero-shot on the seven target benchmarks. As shown in Table~\ref{tab:rho_sens}, $F1_{\text{macro}}$ remains nearly unchanged. However, the learned policy makes slightly more \textsc{Match}/\textsc{Compare} calls and exhibits slightly higher variance across benchmarks.

\section{Confidence Update Rules}
\label{appendix:D}

We maintain per-candidate internal confidence scores $c_i^{(t)} \in [-1,1]$.

\paragraph{\textsc{Match} update.}
Let $s\in[0,1]$ be the matcher probability of [YES]. We map it to $\tilde{s}=2s-1\in[-1,1]$ and update candidate $i$ by exponential smoothing:
\begin{equation}
c_i \leftarrow (1-\alpha)c_i + \alpha \tilde{s},
\end{equation}
where $\alpha=\alpha_{\text{hi}}$ if $|s-0.5|>match\_margin$, otherwise $\alpha=\alpha_{\text{lo}}$.

\paragraph{\textsc{Compare} update.}
Given a compared pair $(i,j)$, the comparator returns a winner and a confidence $q\in[0,1]$. Let $\delta=q-0.5$ and set $\gamma=\gamma_{\text{hi}}$ if $|\delta|>compare\_margin$, else $\gamma=\gamma_{\text{lo}}$. Then:
\begin{equation}
(c_i,c_j)\leftarrow
\begin{cases}
(c_i+\gamma|\delta|,\ c_j-\gamma|\delta|), & \text{if } i\ \text{wins},\\
(c_i-\gamma|\delta|,\ c_j+\gamma|\delta|), & \text{if } j\ \text{wins}.
\end{cases}
\end{equation}

\paragraph{\textsc{Select} update.}
Let $K_t$ be the current top-$k$ shortlist. If the selector outputs a valid index $r\in K_t$:
\begin{equation}
c_r \leftarrow c_r + b^+,\quad \forall j\in K_t\setminus\{r\}: c_j \leftarrow c_j - b^-.
\end{equation}
If the selector output is invalid (e.g., parsing failure), we penalize the shortlist:
\begin{equation}
\forall j\in K_t:\ c_j \leftarrow c_j - b^{\text{inv}}.
\end{equation}
Finally, all scores are clipped to $[-1,1]$.

\section{H100 pricing from different cloud service platforms}
\label{appendix:E}
Table~\ref{tab:accents} lists the H100 GPU server quotes we collected from 4 common cloud service providers. According to this, we estimated the average price of H100 servers on the market.

\begin{table}
  \centering
  \small
  \begin{tabular}{p{1.3cm} p{1.9cm} p{1.6cm} p{1.2cm}}
    \hline
    \textbf{Provider} & \textbf{Instance} & \textbf{GPUs per instance} & \textbf{USD / GPU*h}\\
    \hline
    AWS & p.548xlarge & 8 & \(\approx\) 3.93 \\
    Google & A3-highgpu-1g & 1 & \(\approx\) 3.00 \\
    Azure & Standard-NC40ads-H100-v5 & 1 & \(\approx\) 6.98 \\
    Oracle & BM.GPU. H100.8 & 8 & 10.00 \\\hline
  \end{tabular}
  \caption{H100 GPU server price 4 \$/h .}
  \label{tab:accents}
\end{table}

\section{Blocking Cost and One-time Offline Cost for CaRL-EM}
\label{appendix:F}
We report the blocking cost per dataset, as shown in Table~\ref{tab:blocking_cost}

\begin{table}[h]
\centering
\small
\begin{tabular}{lcll}
\hline
\textbf{Dataset} & \textbf{Cost (\$)} \\
\hline
\textsc{AG} & $1.2e{-4}$ \\
\textsc{DA} & $5.7e{-4}$ \\
\textsc{DS} & $1.02e{-3}$ \\
\textsc{IM} & $9.4e{-4}$ \\
\textsc{IT} & $4.5e{-4}$ \\
\textsc{TT} & $5.5e{-4}$ \\
\textsc{WA} & $1.38e{-3}$ \\
\hline
\end{tabular}
\caption{Blocking cost per dataset.}
\label{tab:blocking_cost}
\end{table}

We report the token and GPU usage during training on AB, as shown in Table~\ref{tab:train_cost}.

\begin{table}[h]
\centering
\small
\begin{tabular}{lcll}
\hline
\textbf{Item} & \textbf{Time/h} & \textbf{Tokens} & \textbf{Cost (\$)} \\
\hline
\textsc{Match} & - & 335,592 & 0.04 \\
\textsc{Compare} & - & 1,694,532 & 0.18 \\
\textsc{Select} & - & 2,122,750 & 0.32 \\
PPO training (H100) & 2.75 & - & 16.45 \\
\hline
\end{tabular}
\caption{One-time offline cost on AB.}
\label{tab:train_cost}
\end{table}

We evaluate the overall end-to-end cost by breaking it down into three distinct parts: the initial blocking cost to build candidate sets, the model training expense, and the inference cost during deployment. Unlike supervised baselines like \textsc{Ditto}, which require separate training for every target dataset, CaRL-EM only needs to be trained once on the source dataset (AB). The learned controller then transfers zero-shot to new target datasets without any retraining. To provide a clear and transparent comparison, we report these three individual components alongside their total. We report the end-to-end cost, as shown in Table~\ref{tab:e2e_cost}.

\textbf{Training Scalability and Wall-Clock Analysis.} To assess the practical scalability of PPO-based controller adaptation, we compare CaRL-EM and \textsc{Ditto} under different training set sizes on Walmart--Amazon (WA). We construct four subsets from WA$'$, containing 25\%, 50\%, 75\%, and 100\% of the training set, respectively. We fine-tune each method on these subsets, evaluate on the same WA test set used in the main paper, and record wall-clock time. We report fine-tuning rather than training from scratch because it is the more practical adaptation setting.

Figure~\ref{fig:train_scaling_combined} shows that CaRL-EM incurs higher wall-clock fine-tuning time, mainly because rollout collection repeatedly invokes backend operators. However, its downstream performance remains relatively stable across training sizes. By contrast, \textsc{Ditto} is much faster to fine-tune, but its performance degrades substantially in the low-data regime. In our setting, the controller itself is not the computational bottleneck: when the LLM operators are replaced with a simulator backend, tuning at the same scale (100k steps) takes only about 1.6 minutes, indicating that backend inference latency, rather than PPO optimization itself, dominates the overall wall-clock cost.

\end{document}